\pdfoutput=1

\documentclass[11pt]{article}

\usepackage{latex/acl}

\usepackage{times}
\usepackage{latexsym}

\usepackage[T1]{fontenc}

\usepackage[utf8]{inputenc}
\usepackage{amsmath}

\usepackage{microtype}
\usepackage{colortbl}
\usepackage{tcolorbox}
\tcbuselibrary{skins}
\tcbuselibrary{raster}
\usepackage{xcolor}
\usepackage{hhline}
\usepackage{inconsolata}
\usepackage{amssymb}
\usepackage{graphicx}

\usepackage{tcolorbox}
\usepackage[frozencache=true,cachedir=minted-cache,outputdir=../]{minted} 
 \definecolor{bananayellow}{rgb}{1.0, 0.88, 0.21}
\usepackage{multicol}

\usepackage{subcaption}
\usepackage{enumitem}
\usepackage{float}
\usepackage{algorithm}
\usepackage{tcolorbox}
\usepackage{algpseudocode}
\usepackage{verbatim}
\newcommand{\TODO}[1]{$ $\newline\noindent\colorbox{yellow!30}{\parbox{\dimexpr\the\columnwidth-2\fboxsep}{\textbf{\texttt{TODO:}} \textit{#1}}}}
\newcommand{\STORY}[1]{$ $\newline\noindent\colorbox{blue!30}{\parbox{\dimexpr\the\columnwidth-2\fboxsep}{\textit{#1}}}}

\title{Interpreting Context Look-ups in Transformers: \\
Investigating Attention-MLP Interactions}

\author{
  Clement Neo$^{\ast \heartsuit \spadesuit}$\quad
  Shay B. Cohen$^{\diamondsuit}$ \quad
  Fazl Barez$^{\ast \heartsuit \clubsuit}$
  \medskip \\
  $^{\heartsuit}$Apart Research \\
  $^{\spadesuit}$Nanyang Technological University \\
  $^{\diamondsuit}$School of Informatics, University of Edinburgh \\
  $^{\clubsuit}$Department of Engineering Sciences, University of Oxford \\
}

\begin{document}
\maketitle
\begin{abstract}
\def\thefootnote{*}\footnotetext{Equal Contribution.}\def\thefootnote{\arabic{footnote}}
Understanding the inner workings of large language models (LLMs) is crucial for advancing their theoretical foundations and real-world applications. While the attention mechanism and multi-layer perceptrons (MLPs) have been studied independently, their interactions remain largely unexplored. This study investigates how attention heads and next-token neurons interact in LLMs to predict new words. We propose a methodology to identify next-token neurons, find prompts that highly activate them, and determine the upstream attention heads responsible. We then generate and evaluate explanations for the activity of these attention heads in an automated manner. Our findings reveal that some attention heads recognize specific contexts relevant to predicting a token and activate a downstream token-predicting neuron accordingly. This mechanism provides a deeper understanding of how attention heads work with MLP neurons to perform next-token prediction. Our approach offers a foundation for further research into the intricate workings of LLMs and their impact on text generation and understanding.
\end{abstract}

\section{Introduction}
\begin{figure}[ht]
  \includegraphics[width=\linewidth]{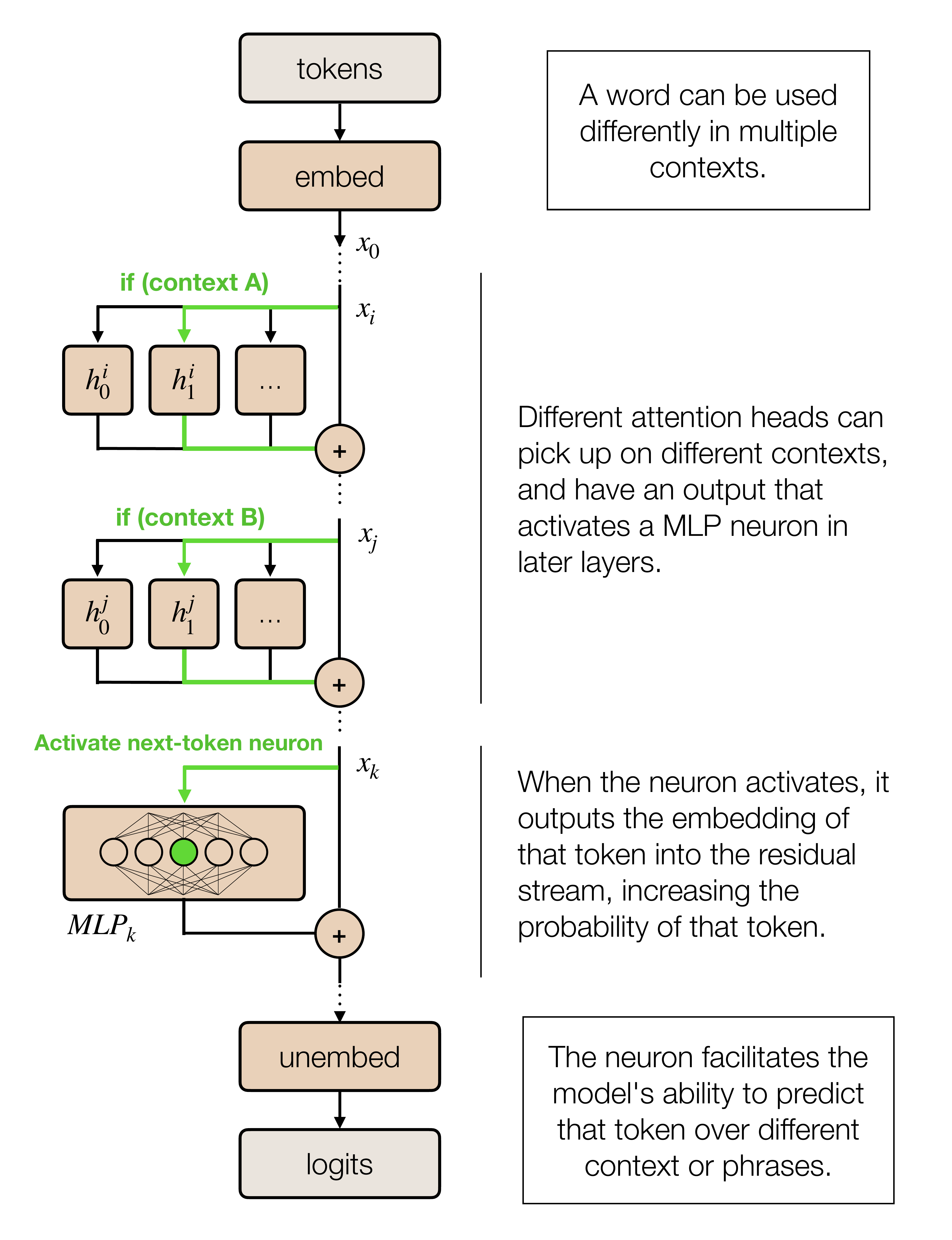}
  \caption{Our approach for characterizing attention heads. For a given token-predicting neuron, we run multiple prompts through GPT-2 and Pythia to find attention heads that activate the neuron. We find that some attention heads activate the neuron only in specific contexts, and use GPT-4 to automate this discovery. }
\label{fig:fig-intro-oa}
\end{figure}

Transformer-based models have made significant advancements in natural language understanding and generation. However, the inner workings of large language models (LLMs) remain largely unexplored, particularly when their behavior deviates from common beliefs \cite{micelibarone2023larger} or when modifying their factual knowledge proves challenging \cite{hoelscherobermaier2023detecting}. A comprehensive understanding of these models is essential for expanding their theoretical foundations, improving real-world applications, mitigating biases, and reducing potential risks.

This study aims to provide insight into the mechanistic interpretability of GPT-2 \citep{gpt2} and Pythia \citep{biderman2023pythia} by examining the associations between their attention heads and multi-layer perceptron (MLP) neurons. Previous research has analyzed attention mechanisms and MLPs separately, demonstrating their ability to model complex input dependencies and generate text \citep{ferrando-etal-2023-explaining}. However, the exact roles, combinations, and interactions of attention heads and MLPs require further investigation.

To address this, we propose a methodology to investigate the interactions between attention heads and next-token neurons. We first identify next-token neurons and find prompts that highly activate them. We then determine the attention heads most responsible for activating each next-token neuron during the forward pass, using a head attribution score. Next, we use GPT-4 to generate explanations for the activity patterns of these attention heads. Finally, we evaluate the explanations' quality by using GPT-4 as a zero-shot classifier to predict head activity on new prompts, comparing the predictions to actual head activity.

Our main contributions are as follows:
\begin{enumerate}
    \item We find that attention heads can recognize contexts relevant to predicting a token, and activate a downstream token-predicting neuron to help predict that token associated with that context;
    \item We develop an automated analysis method to explain and characterize the interactions between attention heads and token-predicting neurons;
    \item We propose an automated approach to evaluate the quality and effectiveness of these explanations.
\end{enumerate}
 
\section{Related Work}
We survey previous work about interpreting language models, with primary focus on attention heads and MLP layers as separate entities. 

\subsection{Attention Interpretability} 
Early work by \citet{elhage2021mathematical} investigated attention-only toy transformers and found that regular transformers use attention-dependent circuits. Building on that work, \citet{wang2023interpretability} uncovered attention mechanisms like Indirect Object Identification (IOI) that copy tokens to predict the next word. However, this mechanism is only relevant when the attention head is copying an existing token, and not predicting a novel word based on the context.

Fully comprehending the functionality of attention heads for more complex tasks has remained elusive. The attention patterns observed cannot be employed as a metric for token saliency \citep{jain2019attention, serrano-smith-2019-attention} due to the distinction between the computational space in which attention heads operate and the vocabulary space, particularly after the initial layers of the model. Hence, we aim to simplify this by associating attention heads with MLP components with known functions.

\subsection{MLP Interpretability} Researchers have investigated the MLP layers at various scales, ranging from the examination of whole layers to groups of neurons and individual neurons.  One such approach involves treating an entire MLP layer as a key-value memory \citep{geva2020transformer}, enabling the modification of factual knowledge in language models \citep{meng2022locating, meng2022memit}. Another approach uses dictionary learning to learn over-complete features of an MLP layer \citep{bricken2023monosemanticity}. Another strategy focuses on identifying meaningful neuron clusters. This has been attempted through the use of linear probes \citep{gurnee2023finding}, and through integrated gradients applied to a cloze task \citep{dai-et-al-2021}. 

A more recent approach focuses on the analysis of individual neurons by inspecting their activation patterns or weights. \citet{bills2023language} and \citet{foote2023neuron} found that certain neurons consistently activate for specific concepts or syntax patterns. To study interactions between attention layers and MLP layers, it would be simpler to consider a neuron as a single analytical unit.

\begin{figure*}[ht!]
  \includegraphics[width=\textwidth]{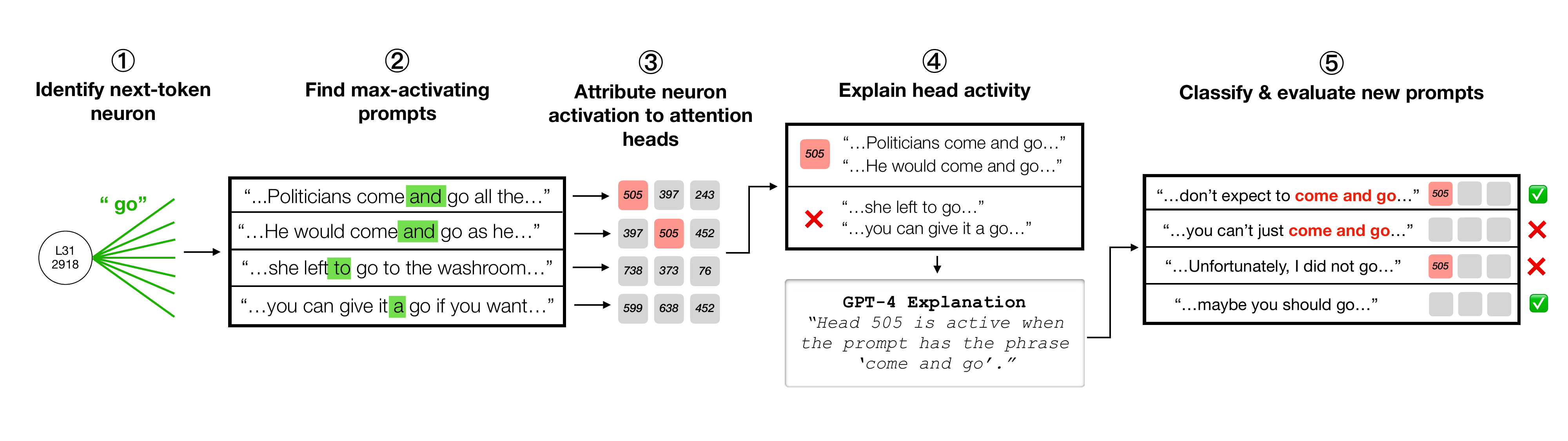}
  \caption{Illustration of our methodology. (1) Identify a token-predicting neuron, characterized by their output weights. (2) Find a set of prompts that highly activate the neuron. (3) Determine the attention heads responsible for activating the neuron during the forward pass for each prompt. (4) Generate explanations for the activity of the attention heads using GPT-4. (5) Use GPT-4 as a zero-shot classifier for test-set prompts using the explanation, based on whether the attention head would be active for that prompt. Evaluate the accuracy of classification. For an example-specific explanation of the illustration, refer to Appendix \S\ref{sec:appA}.}
\end{figure*}

In particular, \citet{geva-etal-2022-transformer} analyzed neuron activation patterns in vocabulary space and identified neurons tied to predicting particular tokens. Building on this, \citet{miller2023anneuron} isolated a neuron specialized for the token ``an'' based on its output weights and prediction correlation. We term these types of neurons ``next-token neurons''.

\subsection{Automated Interpretability}
\citet{bills2023language} introduced the idea of using language models like GPT-4 as tools to explain their own inner workings. They used GPT-4 as an auto-labeler to replace the role of a human labeler to generate explanations at scale. Additionally, they used GPT-4 to assess the quality of these explanations. This approach is promising as it works towards interpretability at scale.

In particular, this approach allows us to examine the interaction between attention heads and next-token neurons, shedding light on how attention focuses processing to predict certain tokens. Our goal is thus to use GPT-4's explanations to better understand attention head behavior in activating specific next-token neurons.

\section{Background}
\textbf{Transformer Models.} We examine transformer-based Language Models (LLMs) characterized by a sequence of length $L$ and dimension $d$, represented as $\mathbf{x}_1, \ldots, \mathbf{x}_L \in \mathbb{R}^d$. Each attention head is defined by parameter matrices $W_k, W_q, W_v \in \mathbb{R}^{d \times d}$. Specifically, for token $i$, we compute $k_i = W_k \mathbf{x}_i$, $q_i = W_q \mathbf{x}_i$, and $v_i = W_v \mathbf{x}_i$. Furthermore, we define $K_i = [k_1, \ldots, k_i] \in \mathbb{R}^{d \times i}$ and $V_i = [v_1, \ldots, v_i] \in \mathbb{R}^{d \times i}$. The output of the head at token $i$ is denoted by $\mathbf{o}_i \in \mathbb{R}^d$, computed as $\mathbf{o}_i = V_i \cdot \text{softmax}(\frac{K_i \cdot q_i}{\sqrt{d_k}})$.

An embedding, represented by function $\Psi : \mathcal{D} \rightarrow \mathbb{R}^d$, maps elements from domain $\mathcal{D}$ to vectors in $\mathbb{R}^d$. Meanwhile, a Multi-Layer Perceptron (MLP) is a neural network layer that processes vectors of size $dl$, transforming them into vectors of the same size. This transformation is achieved through a function $f : \mathbb{R}^{dl} \rightarrow \mathbb{R}^{dl}$, parameterized by weight matrices $W_1$ and $W_2$, and involves applying an activation function $\sigma$. Both the embedding and the MLP layer operate individually on each token \citep{vaswani2017attention}.

\subsection{Relating Attention Heads and Next-Token Neurons}
\label{sec:sub-prob}
Attention heads can be challenging to analyze due to their output in computational space. We simplify this analysis by associating their output to next-token neurons, which are neurons whose output weights correspond with the embedding of a token in the vocabulary. The activation of these neurons correlate with the prediction of their associated tokens.

Hence, we can analyze how individual attention heads in earlier layers contribute to the prediction of a specific token. In a single forward pass, if an attention head's output has a high dot-product with the input weights of a next-token neuron in a later layer, we can say that the attention head is activating that neuron through the residual connection, and hence helping to predict that token.

We can use this interaction to analyze the activity of an attention head in relation to a next-token neuron. For example, if an attention head is highly responsible for activating the ``go'' neuron more frequently in certain prompts compared to others, we can look at the two sets of prompts to find out whether the attention head is responding to specific contexts. In our methodology, we study several models including GPT-2 Large (774M parameters), and use GPT-4 to automate this analysis in an approach similar to \citet{bills2023language}. 

\section{Methodology}
\subsection{Identifying Neurons}
\label{sec:sub_idt}
A next-token neuron refers to an MLP neuron that exhibits a strong association with a specific token. To identify such neurons, we evaluate each neuron's association with all tokens in the vocabulary $\mathcal{V}$ by calculating a congruence score. A neuron with a high congruence score will typically have its associated token as the top prediction when the neuron is highly activating, as shown in Appendix A. The formula for scoring a neuron $i$ is given by:
\begin{equation}
s_{i} = \max_{t \in \mathcal{V}} \langle \mathbf{w}_{\text{out}}^{i}, \mathbf{e}^{t} \rangle,
\end{equation}
\noindent where $\mathbf{w}_{\text{out}}^i$ is the output weights of neuron $i$, $\mathbf{e}^t$ is the embedding vector of a token $t$, and $\langle \cdot, \cdot \rangle$ denotes dot product. 

\begin{figure}
    \includegraphics[width=\linewidth]{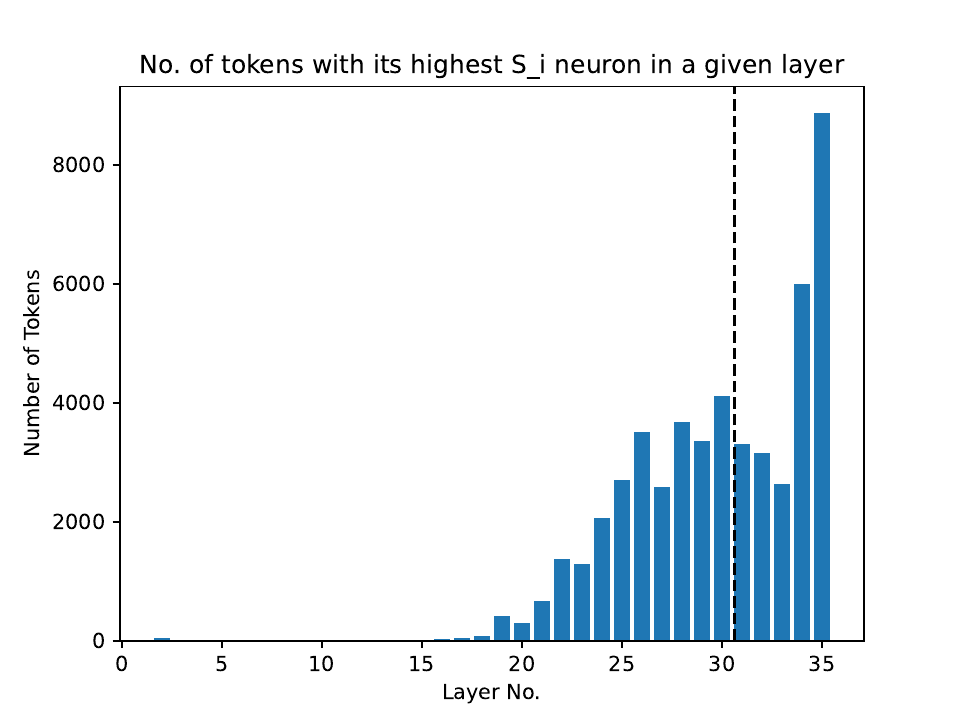}
    
    \caption{Distribution of the set of neurons with the highest score for each token for GPT-2 Large. For each token, we find its highest scoring neuron. We find that these highest-scoring neurons tend to be in the later layers, and we take the last five layers for GPT-2 large (right of dotted line).}
    \label{fig:neuron_over_layers}
\end{figure}

In our analysis, we focus on the top 40 neurons with the highest scores, $s_i$, for each of the last five layers of the model. We select these layers as they have been shown to have neurons that are the most closely aligned with the vocabulary space, as evidenced in the patterns displayed by Figure~\ref{fig:neuron_over_layers}.

\subsection{High-Activating Prompts}
\label{sec:sub_high-a-p}
To study these neurons in their max-activating contexts, we define a function $\psi(p)$ that finds the max activation a neuron achieves for any token position within a prompt $p$. To identify these high-activation prompts, we initially run 10,000 prompts through the model. From these, we select the top 20 prompts for each neuron, noted as $P = [p_1, p_2, ..., p_{20}]$, which is sufficient for finding interpretable activity in their associated attention heads. We further discuss the number of prompts used in \S \ref{sec:head-ex-num-prompts}.

\textbf{Prompt pre-processing.} After identifying these top 20 prompts, we refine them for clarity and conciseness by truncating the prompts from the beginning to the point where they retain 80\% of their initial activation, represented by the condition
$\psi(p_i^*) \geq 0.8 \cdot \psi(p_i)$ where $p_i^{*}$ is the shortened prompt. 
We then discard any prompts exceeding 100 tokens in length and replace line breaks with spaces. The method and its impact on prompt clarity is shown in in Figure \ref{fig:prompt_truncation}.

\begin{figure}[t]
    \centering
    \fbox{ 
        \begin{minipage}{0.95\columnwidth} 
            \begin{center}
            \textbf{Original Prompt}
            \end{center}
            
            \vspace{0.2cm} 
            
            \small
            \textit{(Full news article)...hopes to leave a mark on Oakland that will last a century. The new president of the Athletics is taking on what at one point seemed impossible: building a new ballpark for the team in the city. But while baseball stadiums can come and... (go)}
            
            \begin{center}
            \begin{tikzpicture}
                \draw[->, thick] (0,0) -- (0,-0.5); 
            \end{tikzpicture}
            \end{center}
            \normalsize
            \begin{center}
            \textbf{Truncated Prompt}
            \end{center}
            \small
            \vspace{0.2cm} 
            
            \small
            \textit{while baseball stadiums can come and... (go)}
        \end{minipage}
    }
    \caption{An example of prompt truncation. The activation of the " go" neuron is a minimum of 80\% as compared to its activation for the original prompt. This pre-processing step automatically shortens the prompt while still making sure it activates the neuron significantly.}
    \label{fig:prompt_truncation}
\end{figure}

\textbf{Dataset source.} The 10,000 prompts are from The Pile \citep{pile}\footnote{Released under the MIT license.}, a diverse, open-source dataset composed of several smaller datasets of different domains, including code, books, and academic writing. By using the Pile, we can identify prompts in which next-token neurons activate over different contexts. This helps to ensure that the model behavior is sufficiently general across different contexts. The dataset is further
discussed in \S \ref{sec:interp_illusion}.

\subsection{Individual Head Attribution}
\label{sec:sub-indi-a-h}
In the forward pass of a given prompt, we analyze the associations between attention heads and a next-token neuron by looking at how much each attention head activates the neuron. Formally, for a given attention head and neuron pair, we calculate their head attribution score as: 
\begin{equation}
A_{(i,k), (j,\ell)} = 
\begin{cases} 
\langle h_{i,k} , e_{j,\ell} \rangle, & \text{if } k \leq \ell \\
0, & \text{otherwise,}
\end{cases}
\end{equation}
\noindent where $A_{(i,k), (j,\ell)}$ is the head attribution score of an attention head $i$ at layer $k$ for neuron $j$ at layer $\ell$, $h_{i,k}$ is the attention head's output, and $e_{j,\ell}$ is the neuron's input weight.

To find the attention heads that significantly activate the neuron, we find the attention heads whose head attribution score is $2\sigma$ higher than the mean head attribution score. Hence, for a given prompt, we refer to such attention heads with a high attribution score as \textbf{``active''}.

\subsection{Explaining Attention Heads}
\label{sec:sub-exp-a-h}
We take the set of attention heads that are active for 25\% to 75\% of the max activating prompts for a given neuron. For a given attention head, we categorize the prompts by head activity. With these two classes of prompts, we use GPT-4 to generate an explanation for the attention head (the full prompt can be found in Appendix \ref{sec:appB}). If the attention head is indeed active in specific contexts (e.g. certain phrases), GPT-4 can generate a good explanation for this activity.

To evaluate the quality of these explanations, we find a new set of 10 max-activating prompts on a separate subset of The Pile in the same manner as described in \S\ref{sec:sub_high-a-p}. We then use GPT-4 to classify whether a given attention head would be active for a prompt, given only the explanation (The full prompt can be found in Appendix \ref{sec:appB}). We can then evaluate whether these prompts activate the head as expected, according to whether the head is active as described in \S\ref{sec:sub-indi-a-h}. If the explanation is faithful to the attention head’s activity, GPT-4 should be able to correctly predict whether the attention head is active for a given prompt. We take the ground truth of whether the attention head is active as described in \S\ref{sec:sub_high-a-p}.

To measure explanation quality, we define the head explanation score of head $i$ at layer $k$, $\mathcal{E}^k_i$ as the average of the true positive rate and true negative rate of the classification results:
\begin{equation}
\mathcal{E}^{k}_i = \frac{1}{2} \left( \frac{\text{TP}}{\text{TP}+\text{FP}} + \frac{\text{TN}}{\text{TN}+\text{FN}} \right)
\end{equation}
given the counts of true positives $\text{TP}$, false positives $\text{FP}$, true negatives $\text{TN}$, and false negatives $\text{FN}$. These counts are derived from comparing GPT-4's classifications against the ground truth of head activity in our test set of 10 prompts, where positive examples are prompts with active heads and negative examples are those with inactive heads. If there is no classification made for one class, and either the true positive rate or true negative rate is not applicable, we assign the score to be the other. We also remove cases where all test prompts fall under one classes, as classification can succeed trivially.

The head explanation score is a measure of explanation quality as GPT-4 is performing zero-shot classification given only the explanation. Hence, the higher the head explanation score, the better the explanation is at predicting the attention head's activity.

\section{Results and Discussion}
We first identify meaningful attention head activity (\S\ref{sec:sub_attn_meaningful}), and perform baseline comparisons (\S\ref{sec:baselines}). We then show that this can also be found in smaller model variants (\S\ref{sec:exp_scales}), We verify the relationship between the attention head and the neuron through ablation in  ~\S\ref{sec:head_ablation}. We also discuss the interpretability illusion in this context (\S\ref{sec:interp_illusion}) before discussing the performance of GPT-4 as an auto-labeler in these experiments (\S\ref{sec:gpt4_perf}).

\subsection{Attention Heads May Capture Phrases or Context}
\label{sec:sub_attn_meaningful}

Our analysis reveals that certain attention heads in GPT-2 exhibit consistent activity patterns for prompts that maximally activate specific next-token neurons. For instance, we find heads that activate the "as"-predicting neuron only in contexts like "as early... as" or "as high... as", but not other prompts without these phrases (see Table \ref{tab:prompt_explanation}). When clear differences exist between activating and non-activating prompts, GPT-4 can often generate good explanations of the head's specialized function.

\begin{table}[ht]
\centering
\setlength{\arrayrulewidth}{1.1pt}
\begin{tabular}{l}
\hline
\textbf{Active.} ``...will be available \textit{as early as}...''\\
\textbf{Active.} ``...would reach \textit{as high as}..''\\
\textit{Inactive.} ``...other kinds of work, such as...''\\
\textit{Inactive.} ``...whatever he wants so long as...''\\\hline
\textsc{\textcolor{blue}{Explanation}.} Phrases indicating an\\ approximate point in time, a potential range, or \\a comparison with a certain speed or level.\\\hline
\end{tabular}
\caption{Example of an attention head with semantically meaningful prompt distinctions. This attention head is active when the word ``as'' is predicted in phrases with the form ``as \textit{<\textcolor{blue}{adjective}>}... as''.}
\label{tab:prompt_explanation}
\end{table}

\begin{figure*}[ht]
    \centering
    \begin{subfigure}{0.32\linewidth}
        \includegraphics[width=\linewidth]{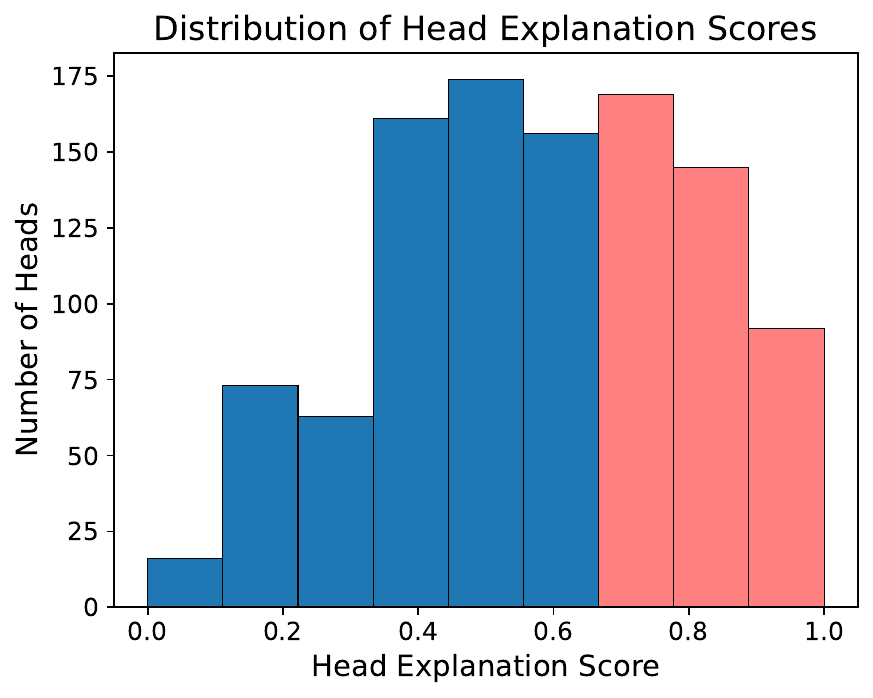}
        \caption{GPT-2 Large\\$\text{Skewness}=-0.132$\\$\text{Mean}=0.575$\\$\text{KS } p\text{-value}=3.2\times10^{-12}$}
        \label{fig:head-ex-group-a}
        \centering
    \end{subfigure}
    \begin{subfigure}{0.32\linewidth}
        \includegraphics[width=\linewidth]{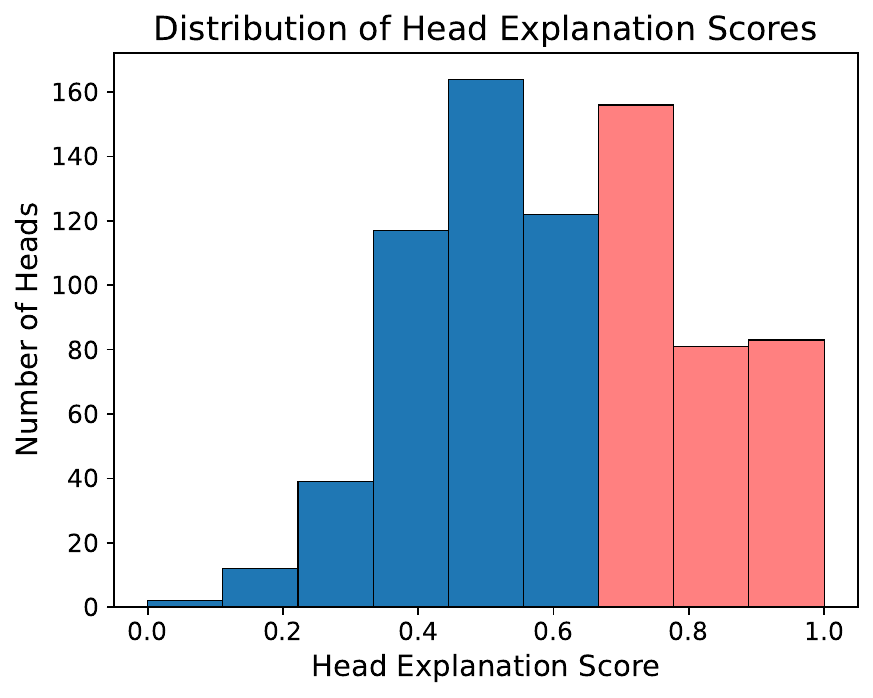}
        \caption{GPT-2 Medium\\$\text{Skewness}=+0.113$\\$\text{Mean}=0.610$\\$\text{KS } p\text{-value}=4.13\times10^{-13}$}
        \label{fig:head-ex-group-b}
        \centering
    \end{subfigure}
    \begin{subfigure}{0.32\linewidth}
        \includegraphics[width=\linewidth]{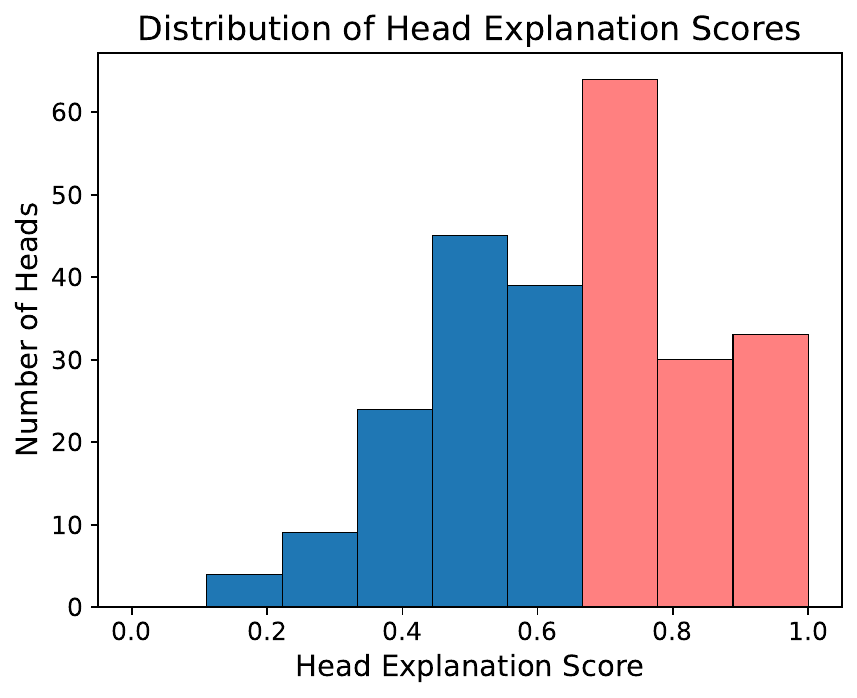}
        \caption{GPT-2 Small\\$\text{Skewness}=-0.132$\\$\text{Mean}=0.650$\\$\text{KS } p\text{-value}=1.6\times10^{-14}$}
        \label{fig:head-ex-group-c}
        \centering
    \end{subfigure}
     \begin{subfigure}{0.32\linewidth}
        \includegraphics[width=\linewidth]{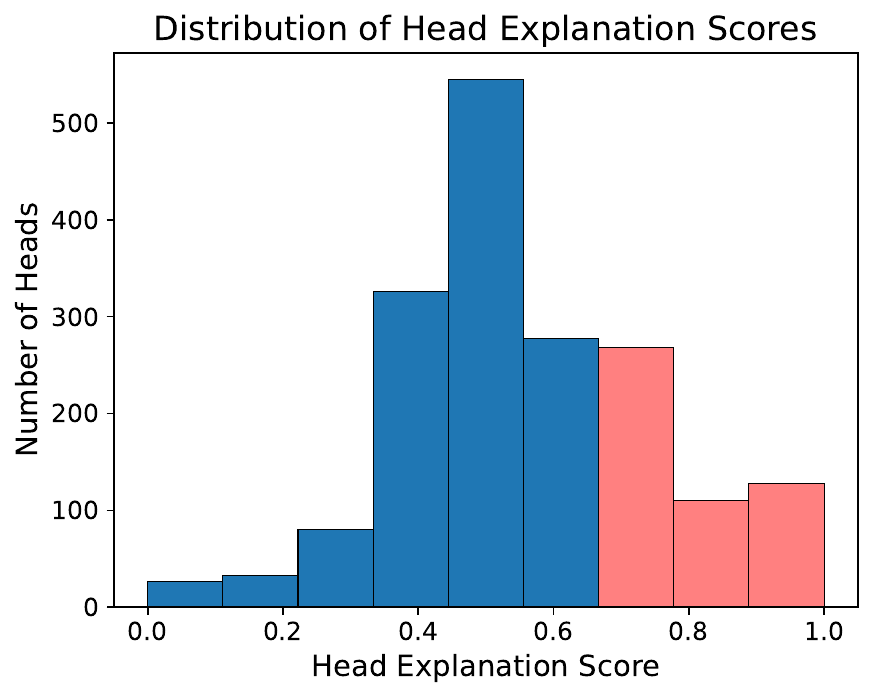}
        \caption{GPT-2 Large (Random)\\$\text{Skewness}=+0.230$\\$\text{Mean}=0.558$}
        \label{fig:head-ex-group-d}
        \centering
    \end{subfigure}
    \begin{subfigure}{0.32\linewidth}
        \includegraphics[width=\linewidth]{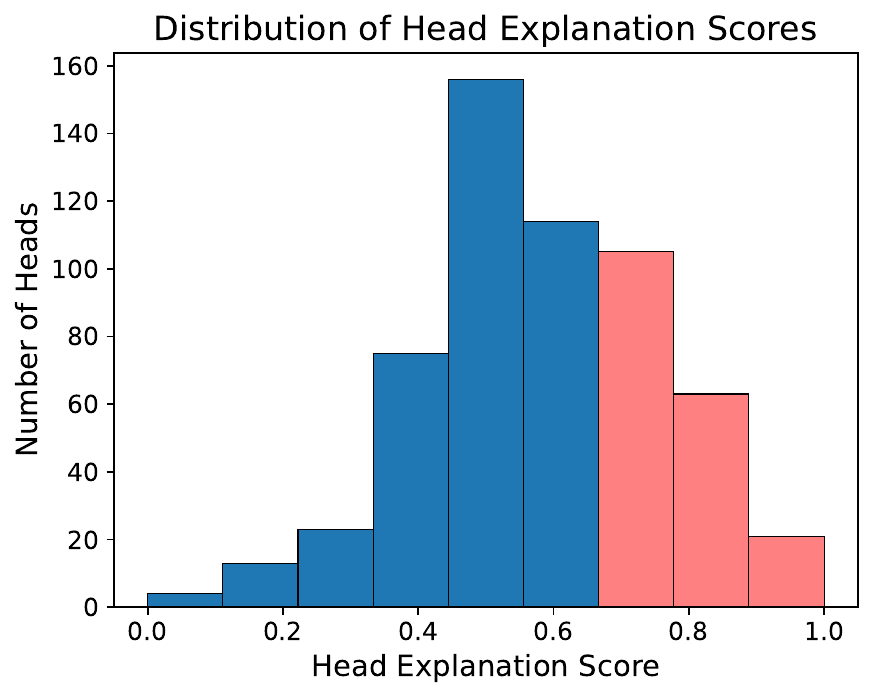}
        \caption{Pythia 1.4B\\$\text{Skewness}=-0.098$\\$\text{Mean}=0.574$\\$\text{KS } p\text{-value}=8.6\times10^{-7}$}
        \label{fig:head-ex-group-e}
        \centering
    \end{subfigure}
    \begin{subfigure}{0.32\linewidth}
        \includegraphics[width=\linewidth]{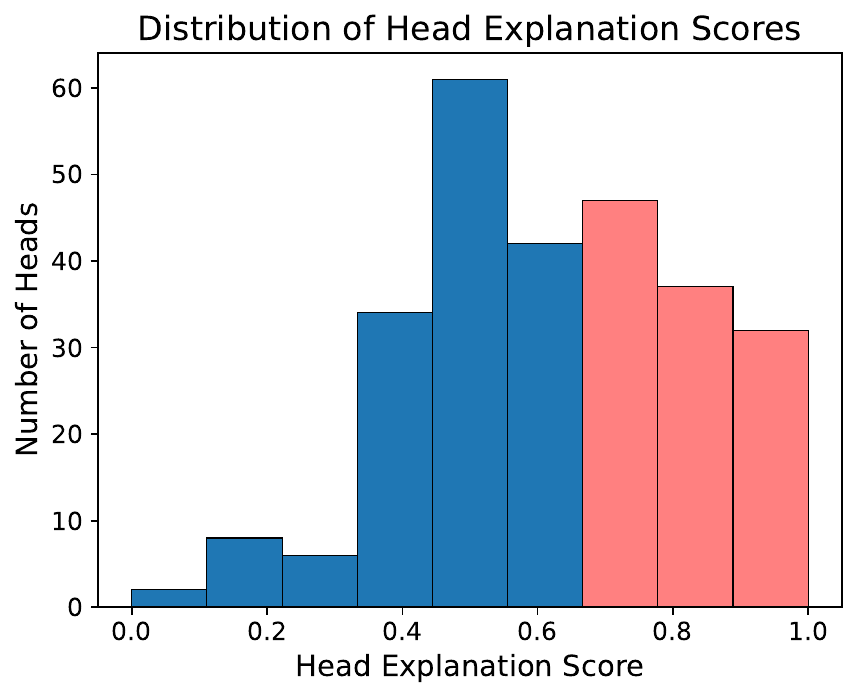}
        \caption{Pythia 160M\\$\text{Skewness}=-0.038$\\$\text{Mean}=0.621$\\$\text{KS } p\text{-value}=1.6\times10^{-7}$}
        \label{fig:head-ex-group-f}
        \centering
    \end{subfigure}
    \caption{The distribution of head explanation scores for GPT-2 and Pythia variants. A negative skew indicates a rightwards-skew with respect to the mean of the distribution, and vice versa.}
    \label{fig:six-graphs}
\end{figure*}

Several heads have high "head explanation scores" (Figure~\ref{fig:six-graphs}), meaning they reliably activate for specific phrases or word usages. Different heads capture various uses of the same word that activate its predicting neuron. For example, the "number" neuron has one head that activates it for ranking concepts (like "number one") and another head for identification numbers ("mobile number", "passport number"). Table \ref{tab:explanation} and  Appendices \ref{app:list}, \ref{app:more_examples} provides examples of explainable heads.


\begin{table*}[!h]
\centering
\begin{tabular}{>{\columncolor{white}}l>{\columncolor{white}}l>{\columncolor{white}}l}
\hhline{===}
\textbf{Neuron-Head} & \textbf{Token} & \textbf{Explanation}\\
\hhline{===}
$(31,3621,538)$ & \verb|"only"| & Describes a specific condition, purpose, or limitation \\
& & \textit{e.g.}, ``for reference purposes only'' or ``by approved personnel only'' \\
\hhline{---}
($31,4378,123$) & \verb|"together"| & When the phrase is ``Taken together'' at the start of a sentence \\
& & \textit{e.g.}, ``...in the brain. Taken together'' \\
\hhline{---}
($31, 364,519$) & \verb|"number"| & When used in the context of ranking or position \\
& & \textit{e.g.}, ``peaked at number one'' or ``debuted at number one'' \\
\hhline{---}
($31, 364,548$) & \verb|"number"| & When referring to an identification, contact, or reference number \\
& & \textit{e.g.}, ``mobile number'', ``account number'', or ``passport number'' \\
\hhline{===}
\end{tabular}
\caption{This table illustrates how specific attention heads activate token-predicting neurons in certain contexts. The notation (Layer, Neuron, Head) is used, where 'Head' is a flattened index. For example, (31, 3621, 538) refers to the 3621st neuron in layer 31, paired with the attention head corresponding to flattened index 538. More examples in Appendix \ref{app:more_examples}.}
\label{tab:explanation}
\end{table*}

Overall, the head explanation score measures explanation quality, since we assume GPT-4 would more accurately classify a prompt given a higher-quality explanation. By examining these scores and prompt activation patterns, we can better understand how GPT-4 represents different linguistic contexts in its attention layers.

\captionsetup[subfigure]{justification=centering} 

\subsection{Baseline Comparisons}
\label{sec:baselines}
\subsubsection{Number of Prompts for Head Explanation}
\label{sec:head-ex-num-prompts}
To evaluate the number of prompts that would be useful for head explanation in \S \ref{sec:sub-exp-a-h}, we performed the methodology using 10, 20 and 30 prompts for the head explanation step for GPT-2 Large. We found that 20 prompts worked the best with the highest mean $\mathcal{E}$ and rightward skew, as shown in Table \ref{tab:head-exp-prompts}.
\begin{table}[H]
\centering
\begin{tabular}{ccc}
\hline
No. Prompts & Mean $\mathcal{E}$ & Skew  \\ \hline
10                & 0.571               & +0.02 \\
20                & 0.575               & -0.13 \\
30                & 0.565               & -0.05 \\ \hline
\end{tabular}
\caption{Comparison of mean $\mathcal{E}$ and skew for different numbers of prompts in head explanation.}
\label{tab:head-exp-prompts}
\end{table}

\subsubsection{Random Late-Layer Neurons}
To determine the importance of choosing next-token neurons, we carried out the methodology on a random set of 20 neurons in each of the last five layers GPT-2 Large.

For these random neurons, head explanation scores skewed less strongly rightward (Figure \ref{fig:head-ex-group-d}). The explanation scores for these layers had a left skew of +0.23. The skew in the other direction indicates explanations were less useful for predicting neuron behavior. 

\subsection{Finding Explainable Heads in Other Models}
\label{sec:exp_scales}
To evaluate whether our findings apply to models of different families and sizes, we applied our methodology to smaller GPT-2 variants, namely GPT-2 Small with 124M parameters and GPT-2 Medium with 355M parameters. We also did this for Pythia variants, namely Pythia 1.4B and Pythia-160M \citep{biderman2023pythia}. We selected the last 4 layers of GPT-2 Medium, Pythia 1.4B, and the last 3 layers of GPT-2 Small, Pythia 160M. In these layers, we selected the top 20 next-token neurons. 

The head explanation score distributions for these models skewed rightwards (Figures \ref{fig:six-graphs}). These rightward skews indicate that the head explanations provided useful information to classify prompts above random chance. However, the presence of skew alone is not sufficient to draw this conclusion.

To further support this finding, we also find that the mean head explanation scores for all models were higher than the random baseline. Additionally, the Kolmogorov-Smirnov test revealed a a statistically significant difference between the score distributions of the model and the random baseline (values in Figure \ref{fig:six-graphs}). These results strengthen the evidence that the head explanations offer valuable information for prompt classification beyond random chance.

\subsection{Verification by Head Ablation} \label{sec:head_ablation}
\begin{figure*}[!h]
    \centering

    \includegraphics[width=0.8\linewidth]{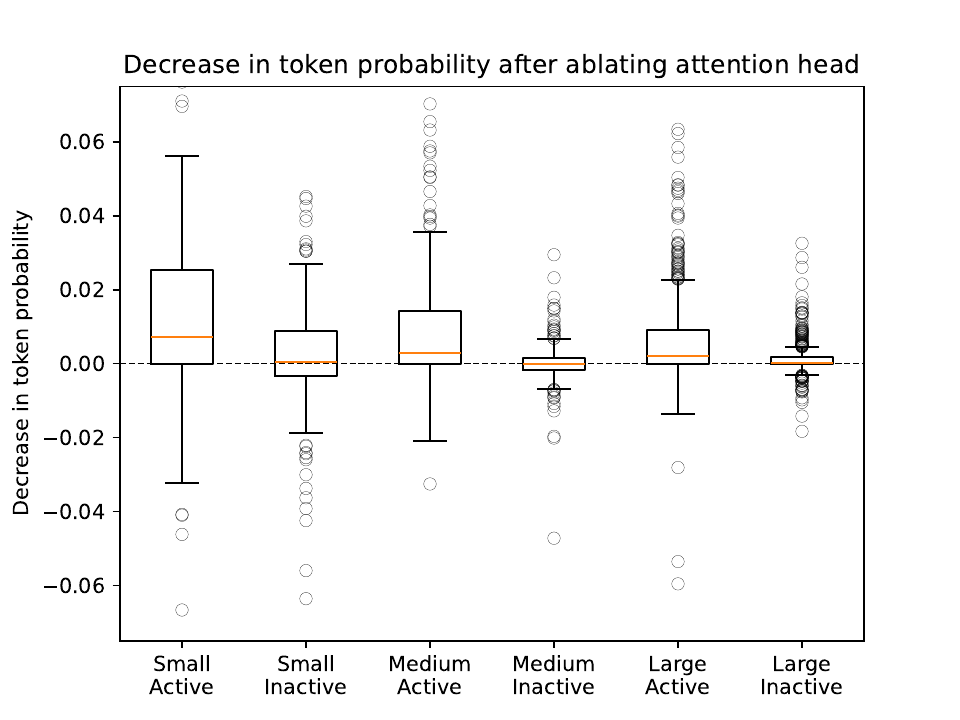}
    \caption{Decrease in token probability after ablating an associated attention head. When the ablated attention head was active for a given prompt, there is a larger decrease in token probability. This applies across model variants. The outliers have been truncated for better visibility.}
    \label{fig:ablate_head_act}
\end{figure*}

To validate the faithfulness of the neural explanations, we ablate selected attention heads when running test prompts through the model. If the explanations fully capture head behavior, the token probability should decrease only on prompts where the head is described as active.
We perform ablation experiments by zeroing out individual heads and comparing token probabilities between original and ablated models on the same prompts. As Figure \ref{fig:ablate_head_act} shows, there is a statistically significant decrease in token probabilities on prompts labeled as head-active versus head-inactive.

Applying a two-sample Kolmogorov-Smirnov test to ascertain the differences in their distributions, we find a significant difference between active and inactive prompts for token probability across all scales, with $p$-values of $p = 1.02\times10^{-5}$, $p = 3.69\times10^{-13}$, $p = 2.09\times10^{-23}$, for the Small, Medium and Large versions respectively.

These targeted ablation results support the hypothesis that the heads help to predict its associated token by firing its next-token neuron. Ablation serves as additional verification that the identified activation patterns are indeed tied to the attention head in question. However, the magnitude of change in token probability is relatively small compared to their original values. This could stem from similar heads active in the same contexts, or downstream effects like backup heads that fill the ablated head's role, as noted by \citet{wang2023interpretability}.

\subsection{Addressing the Interpretability Illusion} \label{sec:interp_illusion}
The ``interpretability illusion'' proposed by \citet{Bolukbasi-et-al-2021} showed neurons can exhibit different activation patterns across datasets, complicating interpretation. However, our approach may avoid this illusion for two key reasons:
First, we specifically focus on next-token neurons, which are more likely to have consistent behavior predicting their associated token, regardless of context. This contrasts with more polysemantic general neurons studied by \citet{Bolukbasi-et-al-2021}.

Second, our experiments leverage The Pile dataset \citep{pile} - a diverse collection of multiple smaller datasets. As Table \ref{tab:pile_counts} illustrates, our prompt sets comprise examples from at least two distinct Pile subsets. This diversity makes consistent neuron behavior more meaningful.
However, the illusion could still occur if a neuron's maximal activation only happened in specific contexts, like solely natural language text. This could result in the top prompts comprising just one Pile subset. Studying neurons outside their max range is outside our scope but merits future work, as discussed in \S\ref{sec:conc}.

\begin{table}[htbp]
    \centering
    \setlength\tabcolsep{5.5pt}
    \begin{tabular}{c|ccccccccc}
        \hline
        \textbf{Datasets} & \textbf{1} & \textbf{2} & \textbf{3} & \textbf{4} & \textbf{5} & \textbf{6} & \textbf{7} & \textbf{8} & \textbf{9} \\
        \hline
        Count & 0 & 4 & 10 & 15 & 12 & 6 & 2 & 0 & 1 \\
        \hline
    \end{tabular}
    \caption{Number of unique datasets that make up each set of max-activating prompts for each neuron-head pair.}
    \label{tab:pile_counts}
\end{table}

\subsection{Strengths and Failures of GPT-4} \label{sec:gpt4_perf}
When inspecting the GPT-4 generated explanation for clear errors, we found a class of attention heads that were active only when the prompt contained a repeated phrase that was about to be followed by the neuron's token. These prompts are similar to the prompts that induction heads work on \citep{olsson2022context}. GPT-4 fails to provide a good explanation for this pattern. Instead, it either offers an overly broad explanation, or attributes the activity to the longer length of the active prompts, which is often an artifact of our prompt truncation. A full example is available in Appendix \ref{sec:appB}.

In general, zero-shot GPT-4 tends to be better in picking up on semantics or specific phrases. This may be because such semantics and phrases are present over multiple active prompts, while for induction-style prompts each prompt has its own unique phrase. 

\section{Conclusion} \label{sec:conc}
We investigated attention heads and their role in activating next-token neurons. We found that some attention heads would only activate their associated next-token neurons in a subset of max-activating prompts. We then use GPT-4 to generate explanations for this activity, and made GPT-4 use that explanation for a classification task on a test-set of prompts to evaluate explanation quality.  Our findings reveal that attention heads process context and use next-token neurons to help predict a word associated with, but not previously seen in that context. This provides insight into how attention heads can work with MLP neurons to predict new words.  

Future work could investigate whether this phenomenon happens to such neurons throughout their range of activations, which could give insights to the role of next-token neurons when the predicted token is unlikely to be the neuron's associated token. Further research could also look into whether multiple attention heads could work with a single next-token neuron in a similar manner, or whether a single attention head could work with multiple next-token neurons in a similar manner. This could provide insights to whether such attention-head neuron interactions extend to more complex contexts.

\section*{Limitations}
We acknowledge several limitations that may impact the analysis and the generalization of the findings. 

First, we focused solely on the max-activating behavior of neurons, which might overlook important nuances or subtler activation patterns since the max-activating range is only a subset of the neuron's full behavior. 

Our investigation relied on a maximum of 30 prompts per neuron. We also observed that GPT-4 struggles to generate coherent and accurate explanations in response to longer prompts, influencing the quality and consistency of explanations provided. To address these problems, we could feed more prompts per neuron into GPT-4, and perform few-shot explanation generation instead of zero-shot. Notably, \citet{bills2023language} used few-shot generation to explain neuron activity. However, this would incur higher financial costs with GPT-4, and it is necessary to strike a balance between thorough research and resource constraints. 

Lastly, the scope of our analysis may not completely capture complex attention head and neuron behavior. We scope only encompassed neurons whose associated tokens are whole words, and we limit our analysis to prompts that are shorter than 100 tokens. Our analysis is a starting point to characterizing such behavior. Therefore, we encourage in future research to employ additional datasets, benchmarking tasks, and examination techniques to enhance our understanding of LLMs strengths and limitations regarding neuron activation, explanation generation, and various aspects of its learned knowledge. 

\section*{Acknowledgements}
We thank Ameya Prabhu, Yftah Ziser, Jason Schreiber for reading drafts of the paper.

\bibliography{custom}

\textbf{AI Assistance.} AI Assistance was used for checks for typos and grammatical errors, as well as suggest alternative phrasings for better expression of existing ideas.

\appendix
\section{Ethical Considerations}
\label{app:sub-app-brd}
This paper investigates the interactions between attention heads and specialized ``next-token'' neurons in the MLP that predict specific tokens. The findings have the potential to improve the efficiency and interpretability of language models, aligning with an overall goal of increasing public trust in AI technologies. 

There is a need for responsible AI research to ensure that novel findings are used for positive societal impacts, while remaining aware of any potential forms of misuse. In interpretability, a better understanding of language models may lead to the dual-use problem and potential misuse. For example, a better understanding of model internals may lead to making the models better at creation of misinformation.  Any progress in understanding language models should be accompanied by an overall understanding of their ethical and safety aspects.

\section{Methodology Example}
\label{sec:appA}
\begin{figure*}[ht]
  \includegraphics[width=\textwidth]{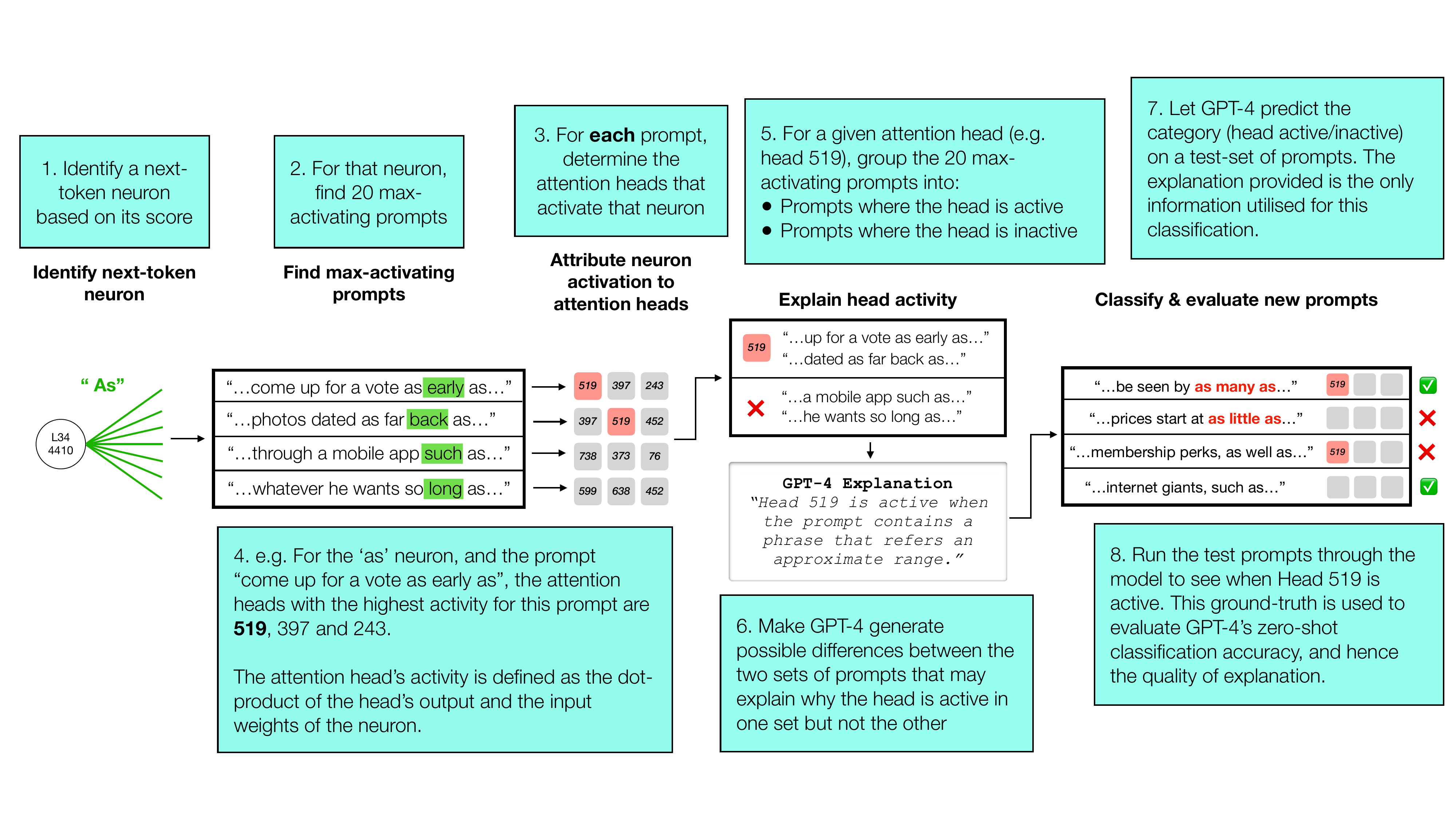}
  \caption{Re-illustration of our method - a step-by-step walkthrough. (1) Identify a token-predicting neuron, characterized by their output weights. (2) Find a set of prompts that highly activate the neuron. (3) Determine the attention heads responsible for activating the neuron during the forward pass for each prompt. (4) Generate explanations for the activity of the attention heads using GPT-4. (5) Use GPT-4 as a zero-shot classifier for test-set prompts using the explanation, based on whether the attention head would be active for that prompt. Evaluate the accuracy of classification.}
  \label{fig:methodology_explained}
\end{figure*}

A walkthrough of our methodology with respect to an example is provided in Figure \ref{fig:methodology_explained}. This example is for the 'as' neuron, where the attention head picks up 'as X as' where it refers to a range of time or quantity)
\section{GPT-4 Prompts}
\label{sec:appB}
Our GPT-4 Prompts used for explanation (Figure \ref{fig:explanation_prompt}), classification (Figure \ref{fig:classification_prompt}), as well as a failure of GPT-4 to explain induction heads (Figure \ref{fig:induction_prompt}).

\begin{figure*}[p]
\centering
\begin{tcolorbox}[colback=white, title={GPT-4 Prompt for Explanation}]
We are studying attention heads in a transformer architecture neural network. Each attention head looks for some particular thing in a short document.

This attention head in particular helps to predict that the last token is `` as'', but it is only active in some documents and not others.

Look at the documents and explain what makes the attention head active, taking into consideration the inactive examples.\\

Examples where the attention head is active:
\begin{itemize}[nosep]
    \item could begin as early as
    \item will be available in schools as early as
    \item The lead would reach as high as
\end{itemize}

Examples where the attention head is inactive:
\begin{itemize}[nosep]
\item  a video game console such as
\item  whatever he wants so long as
\end{itemize}

Explanation: This attention head is active when the document
\end{tcolorbox}
\caption{GPT-4 prompt for generating an explanation for attention head activity based on max-activating prompts it is active and inactive for. This prompt was written in a similar style to the neuron-explanation prompt from \citet{foote2023neuron}. The number of examples in this prompt were shortened for readability (There are typically 15-20 examples total).}
\label{fig:explanation_prompt}
\end{figure*}

\begin{figure*}[p]
\centering
\begin{tcolorbox}[colback=white, title={GPT-4 Prompt for Classification}]
\textcolor{blue}{Is the given example an active example? (Yes/No)}\\

This head is active when the document contains a phrase that describes a specific condition, purpose or limitation before the word ``only.'' In active examples, we see phrases like "for informational purposes only," "for one reason and one reason only," "for general information purposes only," or "for entertainment purposes only." These phrases define a specific context or reason which leads to the appearance of the word "only" at the end. On the other hand, in inactive examples, the word "only" is used in different contexts, without a clear specific condition or purpose preceding it..\\

Example:
\begin{quote}
"All graphs, charts and other visual representations are shown for illustrative purposes only..."
\end{quote}

Answer:
\end{tcolorbox}
\caption{GPT-4 prompt for performing zero-shot classification on a prompt based on the attention head's generated explanation. The second paragraph is the explanation, which is generated by GPT-4 in the previous step.}
\label{fig:classification_prompt}
\end{figure*}

\begin{figure*}[htbp]
\centering
\begin{tcolorbox}[colback=white, title={GPT-4 Failing To Explain Induction-Style Prompts}]
Examples where the attention head is active:
\begin{itemize}[nosep]
    \item out huge amounts of money to go there ... of people willing to shell out huge amounts of money to go there
    \item How many minutes are there between 4:18 PM ... How many minutes are there
    \item accustomed - is there ... time to become accustomed - is there
\end{itemize}
\textcolor{blue}{Explanation}: The head is active when the document
\tcblower
\texttt{
contains a more complex context or multiple ideas and events...}
\end{tcolorbox}
\caption{An incorrect explanation by GPT-4 when given an induction-style set of prompts. The correct explanation would be that these prompts each has a phrase that gets repeated towards the end. The prompts have been simplified for readability.}
\label{fig:induction_prompt}
\end{figure*}

\section{Pseudocode of our Methodology}
\label{sec:appC}
Our methodology is outlined in Algorithm \ref{alg:method}.

\onecolumn

\begin{algorithm}
\caption{Analyzing Attention Head Activity Relative to Next-Token Neurons}
\label{alg:method}
\begin{algorithmic}[1]
\Require GPT-2 model, Set of prompts \( P \), Dataset \( \mathcal{D} \)
\Ensure Explanation scores, Identified active attention heads

\State Initialize GPT-2 model \Comment{\textcolor{blue}{Initialization}}
\State Extract top $n$ next-token neurons from the last $n$ layers \Comment{\textcolor{blue}{Neuron Selection}}

\Function{ProcessPrompts}{neuron}
    \For{each prompt \( p \) in \( P \)}
        \State Score \( p \) based on neuron activation \Comment{\textcolor{blue}{Scoring}}
        \State Shorten \( p \) to retain significant activation \Comment{\textcolor{blue}{Optimization}}
    \EndFor
    \State Return top scored prompts for neuron
\EndFunction

\State \( \text{topPrompts} \) = \Call{ProcessPrompts}{neuron}

\Function{Attribution}{prompt}
    \State Calculate attention head attributions for prompt \Comment{\textcolor{blue}{Attribution}}\\
    \Return dominant attention heads based on attributions
\EndFunction

\State \( \text{dominantHeads} \) = \Call{Attribution}{\text{topPrompts}}

\For{each head in \( \text{dominantHeads} \)}
    \State Generate explanation for head's activity using GPT-4 \Comment{\textcolor{blue}{Explain Activity}}
\EndFor

\For{each new prompt in a subset of \( \mathcal{D} \)}
    \State Classify head activity using the GPT-4 explanation \Comment{\textcolor{blue}{Validation}}
\EndFor

\State Evaluate and report explanation quality \Comment{\textcolor{blue}{Evaluation}}

\end{algorithmic}
\end{algorithm}

\section{Sample of Explainable Attention Head}
\label{app:list}
A sample of how a good explanation for an attention head allows us to classify test prompts correctly is shown in Fig \ref{tab:explainable}.

\begin{figure}[htbp!]
\centering
\begin{tcolorbox}[colback=white]
\textbf{Explainable Attention Heads} \\
When an attention head has a good explanation, it can be used to predict whether the attention head activates its associated neuron for a given prompt. For example, for an attention head associated with a ' way' neuron, if we have the following explanation:\\

\textbf{Explanation.} The attention head in question seems to activate when the phrase "all the way" is used to indicate a complete path, journey, or exhaustive explanation. It is likely looking for contexts where "way" is at the end of a phrase that describes a comprehensive or continuous extent of something. In contrast, the attention head is inactive when "way" is used within idiomatic expressions, comparative phrases (e.g., "the other way around"), or as part of a more metaphorical or abstract concept (e.g., "think of it that way," "put it this way," "stay that way").\\

\textbf{The autolabeller successfully classifies the following examples:}\\

Examples where the head is active:
\begin{itemize}
    \setlength{\itemsep}{0pt}
    \item FROM 0 TO 5, AND THE Y AXIS MUST INCLUDE THE VALUES FROM 236 ALL THE way
    \item stretches as far west as South Dakota, through the northernmost reaches of the U.S. soybean-producing region, all the way
    \item bounding from the Oklahoma panhandle in the 1870s all the way
\end{itemize}

Examples where the head is inactive:
\begin{itemize}
    \setlength{\itemsep}{0pt}
    \item about why things are the way
    \item know G-d's plans/why He arranges things the way
    \item "Wait a minute, it does not work that way
    \item why things are the way
    \item crosses under the New York State Thruway (Interstate 87). Past the Thruway, the commercial buildings give way
\end{itemize}

\end{tcolorbox}
\caption{An example of how an attention head with a good explanation successfully classifies test prompts into active and inactive examples}
\label{tab:explainable}
\end{figure}

\newpage \clearpage
\section{More Examples of Explainable Attention Head Activity}
\label{app:more_examples}

\begingroup
\setlength{\itemsep}{0.2em} 

\subsection{GPT2-Medium}

\begin{itemize}
    \item \textbf{(20, 2312, 225), "day"}
    \begin{itemize}
        \item \textit{Explanation}: Activates with "one day" for future aspirations or developments.
        \item \textit{Examples}: "it may one day", "hoping that will one day"
    \end{itemize}

    \item \textbf{(20, 2312, 66), "day"}
    \begin{itemize}
        \item \textit{Explanation}: Activates with "present-day" or "modern-day" to describe the current state.
        \item \textit{Examples}: "in modern-day", "was the location in present-day"
    \end{itemize}

    \item \textbf{(21, 2629, 228), "right"}
    \begin{itemize}
        \item \textit{Explanation}: Activates with "in the right" implying correct direction.
        \item \textit{Examples}: "point me in the right direction", "to point them in the right"
    \end{itemize}

    \item \textbf{(21, 1031, 20), "off"}
    \begin{itemize}
        \item \textit{Explanation}: Activates with "paid off" or "better off" in contexts of positive outcomes.
        \item \textit{Examples}: "preparation pays off", "hard work and grit always pay off"
    \end{itemize}

    \item \textbf{(21, 3049, 36), "but"}
    \begin{itemize}
        \item \textit{Explanation}: Activates with "no choice but" to indicate compulsion or lack of alternatives.
        \item \textit{Examples}: "have no choice but to", "no option but to"
    \end{itemize}

    \item \textbf{(23, 2257, 59), "some"}
    \begin{itemize}
        \item \textit{Explanation}: Activates with "quite some" indicating an unspecified quantity or duration.
        \item \textit{Examples}: "been wanting to check out for quite some time", "planned feature for inclusion for quite some time"
    \end{itemize}
\end{itemize}

\subsection{GPT2-Small}

\begin{itemize}
    \item \textbf{(10, 2003, 11), "so"}
    \begin{itemize}
        \item \textit{Explanation}: Activates with phrases "or so" or "so" to approximate quantity or degree.
        \item \textit{Examples}: "Another 20,000 or so", "Well, a week or so"
    \end{itemize}

    \item \textbf{(10, 1453, 109), "then"}
    \begin{itemize}
        \item \textit{Explanation}: Activates in shell scripting syntax, particularly with "if" and "then" statements.
        \item \textit{Examples}: "if [ "\$?" -eq "0" ] ; then", "if [[ \$\{PV\} == '9999' ]] ; then"
    \end{itemize}

    \item \textbf{(11, 822, 115), "time"}
    \begin{itemize}
        \item \textit{Explanation}: Activates with "this time" emphasizing the current instance of a recurring action.
        \item \textit{Examples}: "Austin approaches Grandma again, this time", "is back in the news again — this time"
    \end{itemize}

    \item \textbf{(11, 3015, 10), "on"}
    \begin{itemize}
        \item \textit{Explanation}: Activates with "and so on" to imply a non-exhaustive list or continuation.
        \item \textit{Examples}: "input/output devices, and so on", "medical images, demographics, and so on"
    \end{itemize}
\end{itemize}

\subsection{Pythia-160m}

\begin{itemize}
    \item \textbf{(9, 602, 0), "about"}
    \begin{itemize}
        \item \textit{Explanation}: Activates in the context of expressing a gap in knowledge or information.
        \item \textit{Examples}: "very little is known about", "however, little is known about"
    \end{itemize}

    \item \textbf{(9, 2251, 71), "all"}
    \begin{itemize}
        \item \textit{Explanation}: Activates when listing a series followed by "all" as the capper.
        \item \textit{Examples}: "Felipe Anderson, Arthur Masuaku, all", "Hotels.com, Expedia, Hotwire, all"
    \end{itemize}

    \item \textbf{(10, 2509, 80), "more"}
    \begin{itemize}
        \item \textit{Explanation}: Activates when "more" is preceded by "much" at the end of a list.
        \item \textit{Examples}: "Valkyrie Queen, Maggie and the Martians and so much more", "jeans, bags, skirts, and much more"
    \end{itemize}

    \item \textbf{(10, 2734, 10), "on"}
    \begin{itemize}
        \item \textit{Explanation}: Activates with lists that suggest continuation, like "and so on".
        \item \textit{Examples}: "be on the bottom of the stack, and so on", "With one citation in superscript, and so on"
    \end{itemize}
\end{itemize}

\subsection{Pythia-1.4b}

\begin{itemize}
    \item \textbf{(20, 750, 48), "or"}
    \begin{itemize}
        \item \textit{Explanation}: Activates in legal disclaimers, licensing terms, or error handling in code.
        \item \textit{Examples}: "mysqli\_query(\$dbc,\$query) or", "unless required by applicable law or"
    \end{itemize}

    \item \textbf{(20, 3507, 63), "this"}
    \begin{itemize}
        \item \textit{Explanation}: Activates when setting up statements of purpose in research.
        \item \textit{Examples}: "The aim of this study", "The purpose of this investigation"
    \end{itemize}

    \item \textbf{(21, 3152, 290), "at"}
    \begin{itemize}
        \item \textit{Explanation}: Activates in contexts of risk or vulnerability.
        \item \textit{Examples}: "Your capital is at risk", "lives are at stake"
    \end{itemize}

    \item \textbf{(23, 7671, 273), "like"}
    \begin{itemize}
        \item \textit{Explanation}: Activates to introduce examples in comparisons.
        \item \textit{Examples}: "options such as x and y like", "categories such as a and b like"
    \end{itemize}
\end{itemize}
\endgroup

\end{document}